# Beyond Fine-Tuning: Effective Strategies for Mitigating Hallucinations in Large Language Models for Data Analytics

Mikhail Rumiantsau, Aliaksei Vertsel, Ilya Hrytsuk, Isaiah Ballah
team@narrative.bi

## ABSTRACT

Large Language Models (LLMs) have become increasingly important in natural language processing, enabling advanced data analytics through natural language queries. However, these models often generate "hallucinations"—inaccurate or fabricated information—that can undermine their reliability in critical data-driven decision-making. Addressing the challenge of hallucinations is essential to improve the accuracy and trustworthiness of LLMs in processing natural language queries. This research focuses on mitigating hallucinations in LLMs, specifically within the context of data analytics. We introduce and evaluate four targeted strategies: Structured Output Generation, Strict Rules Enforcement, System Prompt Enhancements, and Semantic Layer Integration. Our findings show that these methods are more effective than traditional fine-tuning approaches in reducing hallucinations, offering a more reliable framework for deploying LLMs in natural language queries for data analytics. This research demonstrates the potential of these strategies to enhance the accuracy of LLM-driven data queries, ensuring more dependable results in data-driven environments.

## 1. INTRODUCTION

### 1.1. Background

Large Language Models (LLMs) have become pivotal in the field of natural language processing (NLP), offering powerful capabilities to understand and generate human-like text (Naveed et al.). Their ability to process and interpret vast amounts of textual data has made them essential tools in a variety of applications, from conversational agents to complex data analysis (Kaddour et al.). As organizations increasingly rely on LLMs for tasks such as data retrieval and interpretation, the ability to interact with these models using natural language queries has significantly enhanced the accessibility and usability of data-driven insights (Zhang et al.).

The growing dependence on LLMs for natural language queries in data retrieval and analysis highlights their importance in decision-making processes across industries. By transforming natural language queries into meaningful data outputs, LLMs have the potential to democratize data analytics, allowing users with varying levels of technical expertise to extract valuable information from large datasets. However, as the use of LLMs expands into critical domains, the reliability of these models becomes paramount, particularly in ensuring that the information they generate is accurate and trustworthy.

A critical challenge that has emerged with the increased adoption of LLMs is their susceptibility to generating "hallucinations"—content that is inconsistent with real-world facts, user inputs, or the broader context. The growing reliance on LLMs for data retrieval and analysis through natural language queries has heightened the urgency to address this issue, as hallucinations can lead to the propagation of false or misleading information, undermining the reliability and trustworthiness of LLM-powered data analytics.

## 1.2. The Hallucination Problem in LLMs

Hallucinations in LLMs can manifest in various forms, such as generating factually incorrect information, contradicting previous context, or producing content that is entirely fabricated and ungrounded in reality (Bruno et al.). These errors can have severe consequences in data-driven decision-making, leading to suboptimal or even harmful outcomes.

For instance, an LLM might generate a detailed explanation of a marketing data trend that, upon closer inspection, is based on fabricated data dimensions such as non-existing marketing and advertising campaigns or increased traffic on non-existing web pages. In the context of data retrieval and analysis, such hallucinations can have serious consequences, particularly when the information is used to inform critical decisions. The reliability of LLMs is therefore compromised when hallucinations occur, raising concerns about their deployment in high-stakes environments where accuracy is crucial.

## 1.2. Research Objectives

This research paper aims to address the challenge of hallucinations in LLMs, with a specific focus on their application in data analytics. We introduce and evaluate four targeted strategies to mitigate hallucinations:

1. **Structured Output Generation:** Requiring the model to produce code or structured data before delivering natural language answers.

2. **Strict Rules Enforcement:** Imposing clear guidelines for data retrieval and analysis to avoid inaccuracies.
3. **System Prompt Enhancements:** Augmenting system prompts with contextual metadata to better guide the model's responses.
4. **Semantic Layer Integration:** Assigning synonyms and custom rules to inputs, improving the model's understanding of data structures.

By evaluating these approaches, we aim to identify effective techniques that can enhance the reliability and trustworthiness of LLMs in natural language-driven data analytics, ultimately paving the way for more dependable and accurate data-driven decision-making.

## 2. RELATED WORK

### 2.1. Hallucination in Language Models

Hallucinations in LLMs have been the subject of growing research interest, as evidenced by several recent publications (Yadkori et al.). A comprehensive survey provides a thorough overview of the hallucination phenomenon, including a taxonomy of different types of hallucinations and an analysis of the contributing factors (Huang et al.).

One major cause is the model's tendency to overgeneralize from its training data, producing responses that are statistically likely but contextually irrelevant. Another contributing factor is the model's inability to access or retrieve relevant data during inference, leading it to "fill in the gaps" with fabricated content. Types of hallucinations identified in the literature include factual errors, where the model incorrectly states information, and logical inconsistencies, where the generated output contradicts itself or the input data (Xu et al.).

### 2.2. Traditional Mitigation Techniques

Existing approaches to mitigate hallucinations in LLMs have primarily focused on fine-tuning and data augmentation strategies (Gekhman et al.). Fine-tuning involves adapting a pre-trained LLM to a specific task or dataset, and refining the model's parameters to better align with the desired outputs (Fishman and Anadkat). This technique has been shown to improve the model's performance by making it more sensitive to the nuances of the specific domain it is fine-tuned on. However, while fine-tuning can reduce the frequency of hallucinations, it does not entirely eliminate them. The technique may also introduce new biases if the fine-tuning dataset is not representative or if the process overfits the model to the specific training data (Gekhman et al.).

Prompt engineering, another common approach, involves carefully crafting the input prompts to guide the LLM toward generating more accurate and relevant outputs. By providing clear instructions or specific context within the prompt, it is possible to reduce the likelihood of hallucinations. However, prompt engineering is often limited by the model's inherent capabilities and the ambiguity that may still arise in complex or open-ended queries (Anderson et al.).

Hybrid approaches that combine rule-based systems with LLMs have also been explored as a way to enhance the reliability of model outputs. For instance, rule-based systems can provide a structured framework within which the LLM operates, thereby reducing the risk of hallucinations by enforcing strict adherence to predefined rules or logic. In the context of business intelligence, hybrid systems have shown promise by integrating the precision of rule-based methods with the flexibility and adaptability of LLMs (Vertsel and Rumiantsau). However, these systems also face challenges, such as the complexity of integrating different components and the potential for conflicts between rule-based logic and the generative capabilities of LLMs.

## 3. METHODOLOGY

### 3.1. Structured Output Generation

One of the key strategies we explore to mitigate hallucinations in LLMs for data analytics is Structured Output Generation. Structured Output Generation involves compelling the LLM to produce a structured output, such as code or formal data representations, before delivering answers to natural language queries (Tam et al.). This method leverages the model's ability to generate syntactically correct code or structured data, which is then executed or interpreted to produce a final answer. By requiring the model to generate structured outputs first, we enforce a logical consistency in the responses, reducing the likelihood of hallucinations.

To implement structured output formats, specific templates or coding conventions are predefined within the model's prompt. For instance, when responding to a query about data trends, the model might be instructed to first generate SQL queries or Python code snippets that retrieve and process the relevant data. The structured output is then evaluated for correctness before the model is allowed to produce a natural language response. This two-step process ensures that the final answer is grounded in a verifiable data retrieval process, thereby minimizing errors.

The primary benefit of Structured Output Generation is its ability to reduce ambiguity in the model's responses. By forcing the model to articulate its reasoning in a structured format, we limit the scope for speculative assertions that often lead to hallucinations. This approach also enhances the accuracy of the answers, as the model is guided by a clear logical framework that must be adhered to before producing a response. Moreover, the structured output can be independently verified, offering an additional layer of assurance.

### 3.2. Strict Rules Enforcement

Another approach we investigate is Strict Rules Enforcement, where we impose clear guidelines and constraints on the model's behavior during data retrieval and analysis. A key aspect of this approach is the definition of clear criteria for when the model should abstain from answering (Tam et al.). For example, if the model detects that the available data is insufficient to answer a query accurately, it is programmed to either seek clarification or decline to provide an answer.

The rules are typically formulated based on domain knowledge and the specific requirements of the data being analyzed. For instance, rules might dictate that certain operations, such as statistical analysis or data aggregation, can only be performed if the dataset meets predefined criteria for completeness and relevance. By enforcing these rules, we reduce the likelihood that the model will generate speculative or incorrect outputs.

Implementing Strict Rules Enforcement within the LLM framework involves integrating these rules directly into the model's decision-making process. This can be achieved by embedding the rules in the model's prompt or by using external scripts that evaluate the model's outputs against the rules. For example, if the model is tasked with analyzing sales data, the rules might require that all conclusions be based on a minimum threshold of data points or specific time periods.

### 3.3. System Prompt Enhancements

We also explore System Prompt Enhancements, where we augment the system prompt with additional contextual information and metadata to better guide the model's responses.

System Prompt Enhancements involve the inclusion of contextual metadata within the prompts provided to the LLM (Sahoo et al.). This metadata provides additional context about the dataset, guiding the model's understanding and interpretation of the data. For example, metadata might include information about the data's source, the time period it covers, or specific variables that are relevant to the query. By enriching the prompt with this context, we help the model to better comprehend the nuances of the data, which in turn reduces the likelihood of hallucinations.

The inclusion of contextual metadata can be implemented by appending detailed descriptions of the dataset to the prompt or by embedding structured metadata tags within the input. By providing the model with richer context, we enable it to generate more accurate and relevant responses, as it can draw on the additional information provided to avoid misinterpretations. Enhanced prompts also contribute to more consistent outputs, as the model is less likely to deviate from the intended context when generating its responses. This approach is particularly effective in complex data analysis tasks where understanding the context of the data is crucial for accurate interpretation.

### 3.4. Semantic Layer Integration

Finally, we investigate the integration of a semantic layer to improve the model's understanding of data structures and relationships, aiming to reduce hallucinations by providing more targeted and accurate responses.

We use the semantic layer to assign synonyms and custom rules to inputs, enhancing the model's understanding of the data structure and the relationships between different data elements. The semantic layer acts as an intermediary between the natural language query and the data, translating the query into terms that the model can more accurately interpret. For example, synonyms might be used to standardize the terminology used in queries, ensuring that the model recognizes different expressions of the same concept.

The primary goal of Semantic Layer Integration is to enhance the model's understanding of the data and its underlying semantics (Patel et al.). By providing a structured framework for interpreting natural language inputs, the semantic layer helps the model to better grasp the relationships between different data elements and the specific meaning of terms used in queries. This improved understanding reduces the likelihood of hallucinations, as the model is less likely to misinterpret the input or generate outputs that are inconsistent with the data structure.

Techniques for enhancing data understanding through semantic layers include the use of ontology-based mappings, where terms in the query are mapped to specific concepts within an ontology that defines the data structure. This approach ensures that the model's interpretation of the query is grounded in a well-defined conceptual framework, leading to more accurate and reliable outputs. Additionally, by standardizing the language used in queries through synonyms and custom rules, the semantic layer helps to reduce variability in the model's responses, further minimizing the risk of hallucinations.

Custom rules within the semantic layer can further refine the model's interpretation by defining how certain terms or phrases should be understood in the context of the data. These rules might specify, for instance, that certain terms are synonymous with specific data fields or that particular phrases imply a certain type of analysis. By integrating these semantic enhancements, the model's comprehension of the query is improved, leading to more accurate and relevant outputs.

## 4. EXPERIMENTS AND RESULTS

### 4.1 Datasets

For this study, we utilized a proprietary anonymized dataset, which is based on real-world data collected from various data sources[1] and custom datasets derived from real marketing campaigns. The dataset was carefully anonymized to protect sensitive information while maintaining the integrity and relevance of the data for our research.

The datasets encompass a wide range of data types, including structured data such as user engagement metrics, conversion rates, and ad performance data, as well as unstructured data like

---

[1] These sources include Google Analytics 4, Hubspot, Salesforce, Google Search Console, Google Ads, and Meta Ads

search queries and user feedback. The diverse nature of the dataset allowed us to rigorously test the effectiveness of our hallucination mitigation strategies in different contexts. The experiments were conducted in a controlled computing environment using advanced LLMs and scripts to convert data into structured formats, such as JSON and CSV, to facilitate seamless integration with the LLMs.

To evaluate the combined approach, we utilize a large-scale, obfuscated Google Analytics 360 Dataset[2] containing structured data, including marketing campaigns, website analytics, and transactional records. We modified[3] the timestamps and transaction values in the dataset to preserve the integrity and relevancy of the data.

### 4.2 Evaluation Metrics

To evaluate the performance of our proposed methods in mitigating hallucinations, we used a comprehensive set of metrics that focus on both the accuracy of the model's outputs and the reduction of hallucination occurrences.

The key metrics include:

- **Hallucination Rate:** The percentage of outputs containing fabricated or incorrect information (1).
- **Accuracy:** The percentage of correct responses generated by the model, indicating adherence to the actual data (2).
- **Precision:** The percentage of true positive responses (correct and relevant) to the sum of true positives and false positives, reflecting the model's ability to avoid generating irrelevant information (3).
- **Recall:** The ratio of true positive responses to the sum of true positives and false negatives, indicating the model's ability to retrieve all relevant information (4).

$$Hallucination\ Rate = \frac{Total\ number\ of\ outputs}{Number\ of\ hallucinated\ outputs} \times 100\%\ (1)$$

$$Accuracy = \frac{Total\ number\ of\ responses}{Number\ of\ correct\ responses} \times 100\%\ (2)$$

$$Precision = \frac{True\ Positives}{False\ Positives + True\ Positives} \times 100\%\ (3)$$

$$Recall = \frac{True\ Positives}{True\ Positives + False\ Negatives} \times 100\%\ (4).$$

These metrics provided a robust framework for comparing the effectiveness of our methods against both fine-tuned models and baseline models.

---

[2] https://console.cloud.google.com/marketplace/product/obfuscated-ga360-data/obfuscated-ga360-data
[3] https://github.com/micrum/hallucination-vaccination

## 4.3 Comparative Analysis

### *4.3.1 Structured Output vs. Baseline Model*

In this test, we evaluated the impact of Structured Output Generation on reducing hallucinations compared to a baseline model[4]. The structured output method demonstrated a significant reduction in hallucination rates, particularly in scenarios involving complex data queries.

**Table 1**. Hallucination rate comparison between Structured Output Generation and Baseline model

|  | **Hallucinations: Structured Output** | **Hallucinations: Baseline** |
|---|---|---|
| Data aggregation | 3.1% | 13% |
| Calculated metrics | 1.4% | 21% |
| Data comparison | 1.1% | 4.8% |
| Relational operations[5] | 1.8% | 15.6% |
| Querying against large datasets[6] | 2.1% | 25.3% |
| Table formatted output | 1.1% | 6.7% |
| Chart formatted output | 3.4% | 9.9% |
| Reasoning | 3.8% | 10.2% |

---

[4] GPT-4 Omni
[5] e.g. joins
[6] Datasets over 500MB

*4.3.2 Strict Rules vs. Fine-Tuning vs. Baseline Model*

The Strict Rules Enforcement approach was tested to assess its effectiveness in minimizing hallucinations. Compared to a fine-tuned and baseline model, this method showed a substantial decrease in hallucination rates and an increase in accuracy. The use of strict criteria for when the model should abstain from answering significantly reduced the occurrence of speculative outputs, leading to more precise and reliable responses.

**Table 2.** Accuracy rate comparison between Strict Rules Enforcement, Fine-Tuning, and Baseline model

|  | **Strict Rules, Accuracy** | **Fine-Tuning, Accuracy** | **Baseline model, Accuracy** |
|---|---|---|---|
| Security breach | 98.1% | 97.8% | 91.1% |
| Broken links | 97.3% | 89.1% | 36.1% |
| Forbidden operations | 96.1% | 91.7% | 85.1% |
| Prompt intrusion | 98.5% | 98.1% | 31.7% |

*4.3.3 System Prompt Enhancements vs. Fine-Tuning vs. Baseline Model*

System Prompt Enhancements were evaluated by incorporating contextual metadata within the prompts and comparing the results against fine-tuned and baseline model[7]. The prompts were enriched with contextual metadata that included information about the data's source, the time period it covers, and specific variables that are relevant to the query.

This method resulted in a notable improvement in the model's understanding of queries, reducing hallucination rates and enhancing recall. The enriched prompts guided the model more effectively, leading to more accurate and contextually relevant responses without the unnecessary introduction of new information.

---

[7] Grok-2 Beta

**Table 3.** Comparison of Hallucination rate between System Prompt Enhancements, Fine-Tuning, and Baseline models

|  | Hallucination Rate: Enhanced Prompts | Hallucination Rate: Fine-Tuning | Hallucination Rate: Baseline Model |
|---|---|---|---|
| Unnecessary introductions | 1.2% | 1.5% | 33.9% |
| Generic answers | 2.9% | 8.9% | 20.3% |
| Relative dates understanding | 1.9% | 10.3% | 18.4% |
| Data summarization | 4.8% | 5.0% | 6.5% |
| Visualization type selection | 7.5% | 7.9% | 38.3% |

*4.3.4 Semantic Layer vs. Fine-Tuning vs. Baseline Model*

The Semantic Layer Integration approach was tested to determine its impact on the model's ability to process and understand natural language queries. By integrating synonyms and custom rules, this method improved the model's comprehension of the data structure, leading to a reduction in hallucinations and higher accuracy.

This method works by creating a semantic layer that standardizes language variations and defines custom rules to handle context-specific terms within the data. By aligning various terms and phrases with their underlying data fields and relationships, the semantic layer enables the model to interpret user queries with greater precision. This structured mapping of language to data concepts allows the model to generate responses that are consistent with the dataset's actual content, reducing the likelihood of misinterpretations or hallucinations in responses.

The comparative analysis showed that the semantic layer method outperformed the fine-tuning and baseline model.

**Table 4.** Comparison of Hallucination rate between Semantic Layer Integration, Fine-Tuning, and Baseline models

|  | Hallucination Rate: Semantic Layer | Hallucination Rate: Fine-Tuning | Hallucination Rate: Baseline model |
|---|---|---|---|
| Ambiguous metric names | 1.9% | 7.5% | 10.9% |
| Use of metric synonyms | 2.7% | 4.9% | 11.3% |
| Indirect questions | 1.9% | 10.3% | 18.4% |
| Calculated metrics | 7.5% | 27.9% | 38.9% |
| Reasoning | 8.4% | 21.9% | 48.3% |

*4.3.5 Baseline Models vs. Combined Strategies*

The final experiment assessed the combined impact of the proposed hallucination mitigation strategies, including Structured Output Generation, Strict Rules Enforcement, System Prompt Enhancements, and Semantic Layer Integration. By incorporating these methods simultaneously, model[8] was guided through multiple layers of control and contextual understanding, each method complementing the others to ensure a more robust and accurate response generation.

This comprehensive approach demonstrated the most significant reduction in hallucination rates and the highest overall performance in terms of precision, recall, and accuracy, outperforming the baseline models by a considerable margin.

Together, these combined strategies led to the highest overall performance in terms of precision, recall, and accuracy, significantly outperforming the baseline models by creating a layered mitigation framework that minimized errors and improved the model's reliability across diverse query scenarios.

---

[8] NBI.AI-1

**Table 5.** Performance Comparison: Baseline Models vs. Combined Strategy

|  | **Combined Strategy** | **GPT-4o** | **Grok-2** |
|---|---|---|---|
| Hallucination Rate | 1.52% | 16.67% | 13.64% |
| Precision | 89.39% | 46.97% | 42.42% |
| Recall | 87.88% | 43.94% | 40.91% |

### 4.4. Experimental Setup

To evaluate the effectiveness of the proposed strategies, we conduct a series of experiments on a range of data analytics tasks, including question answering, data summarization, and report generation.

*4.4.1 Data Aggregation Questions*

These questions involve summarizing data across multiple records or metrics.

> **Google Analytics 4 (GA4)**: "What is the total number of users who visited my site in the last quarter?"
>
> **Google Ads**: "How much did we spend on our ads in the last three months across all campaigns?"
>
> **Facebook Ads**: "What is the total number of impressions for my ads in Q3?"
>
> **Google Search Console**: "What is the average number of clicks on my top 10 pages in the last 30 days?"
>
> **HubSpot**: "How many new deals were created this month?"
>
> **Salesforce**: "What is the total value of closed opportunities in Q2?"

*4.4.2 Calculated Metrics Questions*

These questions involve performing calculations based on available metrics.

> **GA4**: "What is the average session duration for users who completed a purchase?"
>
> **Google Ads**: "What is the ROAS for our latest campaign?"
>
> **Facebook Ads**: "What is the click-through rate for my ads targeting mobile users?"
>
> **Google Search Console**: "What is the click-to-impression ratio for branded queries?"
>
> **HubSpot**: "What's the average deal size for deals closed this year?"
>
> **Salesforce**: "What's the average time taken to close a deal across all sales teams?"

*4.4.3 Data Comparison Questions*

These questions require comparing different data sets or time periods.

> **GA4**: "How does the user retention in Q2 compare to Q1?"
>
> **Google Ads**: "How did the conversion rate for mobile users compare to desktop users in the past month?"
>
> **Facebook Ads**: "What's the difference in cost-per-click between our Q3 and Q4 campaigns?"
>
> **Google Search Console**: "How did clicks from organic search perform in September compared to August?"
>
> **HubSpot**: "How does the number of deals closed by the sales team in July compare to June?"
>
> **Salesforce**: "What's the difference in total revenue between Q1 and Q2?"

*4.4.4 Relational Operations (e.g., Joins) Questions*

These questions involve combining multiple datasets or data fields.

> **GA4**: "Which landing pages have the highest conversion rate from users who came via paid search?"
>
> **Google Ads**: "Which campaigns resulted in the highest number of purchases from returning users?"
>
> **Facebook Ads**: "Which ad sets performed best for users who previously interacted with our content?"
>
> **Google Search Console**: "Which pages received the most impressions from users in the USA searching for our brand?"
>
> **HubSpot**: "Which sales representatives closed the most deals in conjunction with marketing-driven leads?"
>
> **Salesforce**: "Which opportunities involved interactions with both the marketing and sales teams?"

*4.4.5 Querying Against Large Datasets Questions*

These questions involve performing operations on extensive datasets[9].

> **GA4**: "Provide user engagement metrics for the 1,000 most-visited pages."
>
> **Google Ads**: "What's the total ad spend for all campaigns targeting users across multiple geographic regions?"
>
> **Facebook Ads**: "How did all the active campaigns targeting users aged 18-24 perform over the past six months?"
>
> **Google Search Console**: "What's the total number of impressions for all the pages indexed in the last year?"
>
> **HubSpot**: "What's the average lifecycle stage for contacts created in 2024"
>
> **Salesforce**: "List all opportunities with a value greater than $100,000 that were created in the last year."

---

[9] Datasets over 500MB

*4.4.6 Table Formatted Output Questions*

These questions ask for results to be presented in a table format.

> **GA4**: "Show a table of sessions and bounce rate by country for last month?"
>
> **Google Ads**: "Create a table comparing cost-per-click, conversion rate, and impressions for all campaigns."
>
> **Facebook Ads**: "Provide a table with the CTR, CPC, and total spend for each ad group in the current campaign."
>
> **Google Search Console**: "Show a table of my top-performing pages with their click-through rate and impressions."
>
> **HubSpot**: "List deal stages and associated deal values for all open deals."
>
> **Salesforce**: "List all open opportunities with their owner, expected close date, and amount."

*4.4.7 Chart Formatted OutputQuestions*

These questions ask for a chart as the output.

> **GA4**: "Generate a chart showing traffic trends over the last 6 months."
>
> **Google Ads**: "Create a bar chart comparing the conversion rates of my top 5 campaigns."
>
> **Facebook Ads**: "Can you show me a line chart of impressions and clicks for my last three campaigns?"
>
> **Google Search Console**: "Display a pie chart of clicks by country for the past 90 days."
>
> **HubSpot**: "Create a clustered stacked bar chart of deal stages by sales representative."
>
> **Salesforce**: "Generate a line chart of opportunity value trends over the last year."

*4.4.8 Reasoning Questions*

These questions involve logical reasoning or conclusions based on the data.

> **GA4**: "What is the likely cause for the significant drop in user engagement last week?"
>
> **Google Ads**: "Why did my conversion rate decrease in the last ad campaign?"
>
> **Facebook Ads**: "Which factors contributed most to the increase in CPC for my mobile ads?"
>
> **Google Search Console**: "What could explain the sudden increase in organic search impressions last month?"
>
> **HubSpot**: "Why did the sales team close fewer deals in August compared to July?"
>
> **Salesforce**: "What might be causing the delay in closing high-value opportunities this quarter?"

**4.5. Discussion**

*4.5.1 Interpretation of Results*

The results of our experiments demonstrate the effectiveness of the proposed methods in mitigating hallucinations in LLMs, particularly when compared to traditional fine-tuning and baseline models.

- **Structured Output Generation** consistently produced the most accurate and logically consistent responses, particularly in complex data processing scenarios.
- **Strict Rules Enforcement** effectively minimized speculative outputs, ensuring that the model only provided responses when there was sufficient data to support them.
- **System Prompt Enhancements** significantly improved the model's contextual understanding, resulting in more relevant and accurate responses.
- **Semantic Layer Integration** enhanced the model's ability to interpret queries accurately, especially in cases involving complex data semantics.
- The **combination of these strategies** outperformed the baseline models across all evaluation metrics, showcasing the synergistic benefits of a multi-pronged approach to hallucination mitigation.

*4.5.2 Comparison with Baseline Models*

Our methods outperformed baseline models across all evaluation metrics, particularly in reducing hallucination rates. The improvements were most pronounced in scenarios involving complex or ambiguous queries, where traditional fine-tuning approaches struggled to maintain accuracy and relevance.

*4.5.3 Potential Limitations and Areas for Improvement*

While the proposed methods demonstrated significant improvements, there are still areas for further research and development. The methods we used may increase computational complexity, particularly in the case of Structured Output Generation and Semantic Layer Integration. Future research could focus on optimizing these methods to balance accuracy with computational efficiency.

## 5. CASE STUDIES

**5.1 Real-World Application: AI Data Analyst**

In this case study, we focus on how our hallucination mitigation strategies were applied in the AI Data Analyst[10], a tool designed to answer natural language queries about data. The tool is used in two key scenarios: datasets with a known structure[11] and arbitrary datasets with unknown structures[12].

*5.1.1 Scenario 1: Known Structure (Google Analytics 4)*

For datasets like Google Analytics 4, users often ask questions such as "What are the key trends in website traffic over the past month?" To ensure accurate answers, we implemented **Structured Output Generation**, which required the model to generate and execute Python code before providing a response. This step anchored the model's answers in concrete data, significantly reducing the incidence of hallucinations.

Additionally, we used **Strict Rules Enforcement** to prevent speculative answers. If the data was incomplete or did not meet predefined criteria, the model would abstain from answering or requesting additional context. **System Prompt Enhancements** further improved accuracy by including metadata about the dataset, such as time ranges and relevant metrics, guiding the model to produce more contextually accurate responses.

---

[10] https://nbi.ai/
[11] Google Analytics 4 dataset
[12] CSV files with marketing data

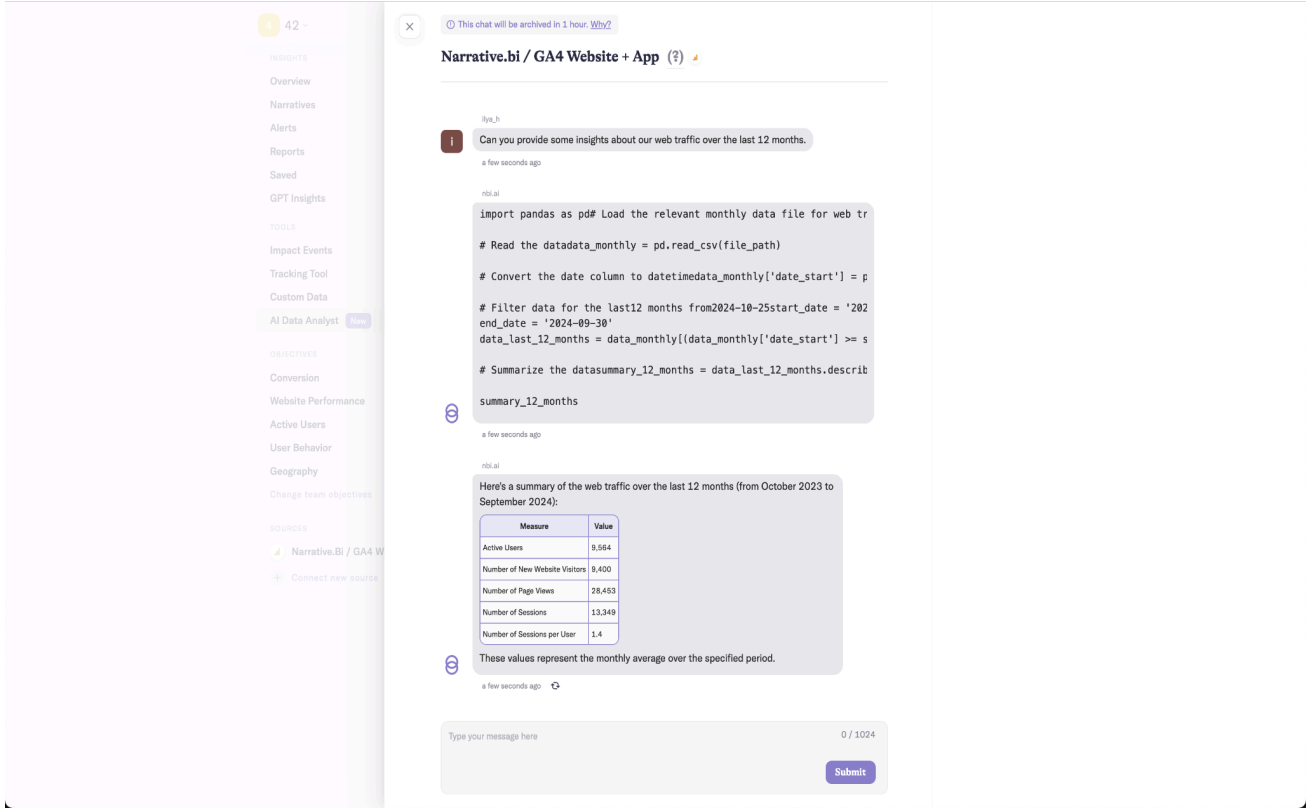

**Figure 1.** Question-answering scenario with known data structure

*5.1.3 Scenario 2: Unknown Structure (Arbitrary CSV Files)*

In scenarios involving arbitrary datasets with unknown structures, such as CSV files from marketing campaigns, users might ask, "Which campaign had the highest ROI?" Here, the **Semantic Layer** Integration played a crucial role. By mapping synonyms and applying custom rules, the model could better interpret the varied and potentially unfamiliar terminology used in these datasets.

For these queries, **Structured Output Generation** helped by first generating code snippets that parsed and analyzed the CSV data before the model attempted to answer the query. This approach ensured that the answers were grounded in the actual data structure, even when the dataset was unfamiliar to the model.

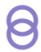

**Figure 2.** Question-answering scenario with an arbitrary dataset

*5.1.4 Results and Impact*

Implementing these strategies resulted in a significant reduction in hallucinations across both scenarios. The AI Data Analyst became more reliable in answering natural language queries, providing users with accurate, actionable insights regardless of the dataset's structure (Vertsel). This case study highlights the importance of tailored hallucination mitigation strategies in enhancing the performance of LLMs in real-world data query applications.

**CONCLUSION**

In this research paper, we have presented a comprehensive investigation of effective strategies for mitigating hallucinations in LLMs used for natural language queries in data analytics. By exploring Structured Output Generation, Strict Rules Enforcement, System Prompt Enhancements, and Semantic Layer Integration, we have demonstrated that these proposed methods significantly outperform fine-tuning in reducing hallucinations.

The superiority of these methods in mitigating hallucinations highlights their potential for broader application in other data-driven tasks where accuracy is paramount. However, there remains room for further exploration. Future research could focus on optimizing these methods to balance computational efficiency with performance.